\documentclass[final]{article}
\usepackage{neurips_2019}
\usepackage{times}
\usepackage{latexsym}
\usepackage{amsmath}
\usepackage{amsfonts}
\usepackage{graphicx}
\usepackage{multirow}
\usepackage{verbatim}
\usepackage{fullpage}
\usepackage{tikz}
\usepackage{hyperref}
\usetikzlibrary{fit,positioning}
\usepackage{comment}
\usepackage{floatrow}
\usepackage{cleveref}
\setcitestyle{numbers}
\newfloatcommand{capbtabbox}{table}[][\FBwidth]

\title{Private Federated Learning with Domain Adaptation}

\author{Daniel W. Peterson \hspace{0.2in} Pallika Kanani \hspace{0.2in}  Virendra J. Marathe\\
  Oracle Labs\\
  \texttt{\{daniel.peterson,pallika.kanani,virendra.marathe\}@oracle.com}
} 

\begin{document}
\maketitle

\section{Introduction}

Federated Learning (FL) is a distributed machine learning (ML) paradigm that enables multiple parties to jointly re-train a shared model
without sharing their data with any other parties~\cite{bonawitz19,konecny15}, offering advantages in both scale and privacy.
We propose a framework to augment this collaborative model-building with per-user domain adaptation.
We show that this technique improves model accuracy for all users, using both real and synthetic data,
and that this improvement is much more pronounced when differential privacy bounds are imposed on
the FL model.

In FL, multiple parties wish to perform essentially the same
task using ML, with a model structure that is agreed upon in advance.
Although the initial focus of FL has been on targeting millions of
mobile devices~\cite{bonawitz19}, the benefits of its architecture
are beneficial even for enterprise settings: the number of users
of an ML service may be much smaller, but privacy concerns are paramount.
Each user wants the best possible classifier for their individual use,
but has a limited budget for labeling their own data. Pooling the data of
multiple users could improve model accuracy, because accuracy generally
increases with increased training data.
The FL framework allows them to effectively pool their labeled
data, without explicitly sharing it. 

Recognizing that just FL is not sufficient to guarantee data privacy, 
many have proposed the addition of differential privacy ~\cite{dwork06,dwork06a,dwork14} to FL~\cite{abadi16,geyer17,konecny16,mcmahan16}.
Informally, differential privacy aims to provide a bound, $\varepsilon$,
on the variation in the model's output based on the inclusion or exclusion
of a single data point. Introducing ``noise'' in
the training process (inputs, parameters, or outputs) makes it difficult
to guarantee whether any particular data point was used to train the model. 
While this noise
ensures \emph{$\varepsilon$-differential privacy}~\cite{dwork06} for the data point,
it can degrade the accuracy of model predictions.

We consider the setting in which individual user data comes from diverse domains.
This is common, because each user can have a different data-generating process.
We show in~\Cref{sec:eval} that, in this setting, the differentially private FL
model may provide worse performance for some users than a non-collaborative baseline,
despite the larger training data set available through FL.
There exists a large body of work on \emph{domain adaptation}
in non-FL systems~\cite{ben-david10,crammer08,kouw18,pan10,daume09}.  In domain adaptation, a model
trained over a data set from a source domain is further refined to
adapt to a data set from a different target domain.

In this work, we use privacy-preserving FL to train a public,
generalist model on the task, and then adapt this general model
to each user's private domain.
While learning the general model, our system also learns a private model for each user.
Each user combines the output of the general and private models using a
mixture of experts (MoE)~\cite{masoudnia14,nowlan1991} to make their final predictions.
The two ``experts'' in the mixture are the general FL model and the domain-tuned private model,
so we refer to our system as federated learning with domain experts (FL+DE).
For privacy in the general model, we use FL with
differentially private stochastic gradient descent (SGD)~\cite{abadi16}.
The private domain models are trained using ordinary stochastic gradient descent (i.e. without
differential privacy noise).  In principle, the two model architectures
can be identical or radically different, but for convenience
we maintain a common model architecture for the general
(public) and private models.  Using a MoE architecture allows the general and
private models to influence predictions differently on each individual data points.

We demonstrate that our system significantly outperforms the accuracy of differentially
private FL. This largely boils down to two
factors. First, the private models provide domain adaptation, which is
known to typically increase accuracy in each domain. On a real-world classification task, we observe a 1.3\% absolute
accuracy improvement due to domain adaptation alone. Second, the private
models allow noise-free updates, because there is no need to conceal
private data from the private model. While the accuracy of the differentially-private FL system degrades by 11.5\% in the
low-noise setting, the performance of FL+DE does not degrade at all. In the high-noise
setting, the accuracy of the differentially-private FL system degrades by 13.9\% and FL+DE accuracy degrades by only 0.8\%.

\section{Our Model}


At its core, our proposal is to mix the outputs of
a collaboratively-learned general model and a domain expert.
Each participating party has an independent set of labeled training examples that
they wish to keep private, drawn from a party-specific domain
distribution. 
These users collaborate
to build a general model for the task, but maintain private, domain-adapted
expert models.
The final predictor is a weighted average of the outputs
from the general and private models.
These weights are learned using a
MoE architecture~\cite{masoudnia14,nowlan1991}, so the entire model can be trained with gradient descent.

More specifically, let $M_G$ be a general model, with parameters
$\Theta_G$, so that $\hat{y}_{G} = M_G(x, \Theta_G)$ is $M_G$'s predicted
probability for the positive class, or perhaps a regressed value\footnote{Although
we tested only binary classification and regression in our experiments, there are obvious extensions to
multiclass problems.}.
$M_G$ is shared between all parties, and
is trained on all data using FL with differentially private SGD~\cite{abadi16}, enabling
each party contribute to training the general model.

Similarly, let $M_{P_i}$ be a private model of party $i$,
parameterized by $\Theta_{P_i}$, and $\hat{y}_{P_i} = M_{P_i}(x,
\Theta_{P_i})$ be the model's predicted probability. Although $M_{P_i}$ could have a different architecture
from $M_G$, in this work we initialize $M_{P_i}$ as an exact copy of
$M_G$. Neither $M_{P_i}$, nor gradient information about it, is shared
with any other party, so $M_{P_i}$ can be updated exactly, without
including privacy-related noise.

The final output that party $i$ uses to label data is
\begin{equation}
  \hat{y}_i = \alpha_i(x) M_G(x, \Theta_G) + (1 - \alpha_i(x)) M_{P_i} ( x, \Theta_{P_i}).
  \label{eq3}
\end{equation}
\vspace{-0.2in}

The weight $\alpha_i(x)$ is called a gating function in the MoE literature. 
In our experiments, we set $\alpha_i(x) = \sigma(w_i^{T} \cdot x + b_i)$,
where $\sigma(x)$ is the sigmoid function, and $w_i$ and $b_i$ are
learned weights.  This gating function learns which regions to trust the private model over the general model,
and allows smooth mixing along the boundary.
Thus the final output
$\hat{y_i}$ depends on learned parameters $\Theta_G$, $\Theta_{P_i}$,
$w_i$, and $b_i$, and all are updated via SGD.

The private model $M_{P_i}$ and weighting mechanism $\alpha_i$ work
together to provide a significant benefit
over differentially private FL. First, by allowing individual domain adapatation,
they boost accuracy. Second, because they allow noise-free updates, they
prevent the accuracy loss associated with more stringent privacy requirements,
which add noise to the general model.

Over time, a user's gating function
$\alpha_i(x)$ learns whether to trust the general model or the private model
more for a given input, and the private model $M_{P_i}$ needs to perform
well on only the subset of points for which the general model fails.
While the general model still benefits from the pooled training data,
it receives weaker updates on these ``private'' data points. This means
users with unusual domains have a smaller effect on the general model,
which may increase its ability to generalize~\cite{ji2018}.
This may also provide increased privacy for the users' data.

\section{Evaluation}
\label{sec:eval}

The main hypothesis of this work is that domain adaptation techniques can
improve accuracy in a federated learning setting, and that this accuracy
improvement holds even when noise is added to protect the privacy
of the gradient updates.
We illustrate the effectiveness of our domain adaptation technique on two
datasets.

\begin{figure}[t]
\includegraphics[width=\textwidth]{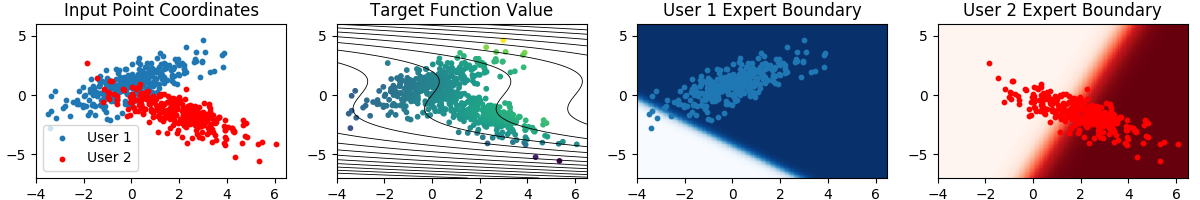}
\caption{Visualizing the synthetic data experiment. Axes for all figures
represent input values for $x_1$ and $x_2$. Left to right:
(a) $x_1$ and $x_2$ values of test data points sampled from distinct 2d Gaussians.
(b) Target values of nonlinear function $f(x_1, x_2)$.
(c) Values of the MoE gating function, $\alpha_1(x_1, x_2)$ learned by User 1. In the darker region,
the private domain expert is preferred, while the general model is preferred in the
lighter region.
(d) The gating function $\alpha_2(x_1, x_2)$ of User 2, which uses the shared model in a
different region than User 1.
}
\label{synthetic_image}
\end{figure}

The first dataset is a synthetic regression problem. Two users attempt
to fit a linear model of the function $f(x_1, x_2) = 5x_1 - 2x_2 + 0.5x_2^3$.
Each has input data drawn from a distinct 2-dimensional Gaussian, and because
of these domain differences, they get different exposure to the nonlinear
$x_2^3$ term. We draw 2500 training examples, 500 validation examples, and 500
test examples for each user, all from that user's 2d Gaussian, then compute $f(x_1, x_2)$.
The users aim to minimize root mean squared error (RMSE) on the
test set. The baseline error is computed with each user fitting a
single linear model to their training data. We then compute RMSE for
each user if the users collaborate to build a single linear model
using FL, and augmenting FL with private domain experts (FL+DE).
Figure~\ref{synthetic_image} shows the synthetic data, the
target function, and the learned gating functions for both users.
To see the effects of differential privacy, we test with low noise ($\sigma=2$)
and high noise ($\sigma=4$), following prior work~\cite{abadi16}.

\begin{figure}
\begin{floatrow}
\capbtabbox{\TopFloatBoxes
\begin{tabular}{|c|c|c|}
\hline
{\bf System} & {\bf User 1 RMSE} & {\bf User 2 RMSE} \\
\hline
Baseline & 15.32 & 10.95 \\
\hline
FL, $\sigma = 0$ & 12.75 & 9.67 \\
FL, $\sigma = 2$ & 13.79 &  12.68 \\
FL, $\sigma = 4$ & 12.59 & 19.49 \\
\hline
FL+DE, $\sigma = 0$ & 12.12 & {\bf 9.41} \\
FL+DE, $\sigma = 2$ & {\bf 12.05} & 9.73 \\
FL+DE, $\sigma = 4$ & 13.78 & 10.95 \\
\hline
\end{tabular}
}
{
\caption{ Test error for regression models trained on synthetic data
  (lower is better).  The domain-only baseline system trains a separate model for
  each user on their data. Traditional FL, and our system of FL with domain experts 
  (FL+DE) are tested with various noise levels, $\sigma$, for differential privacy.}
\label{synthetic-table}
}
\ffigbox{\TopFloatBoxes
\includegraphics[width=\linewidth]{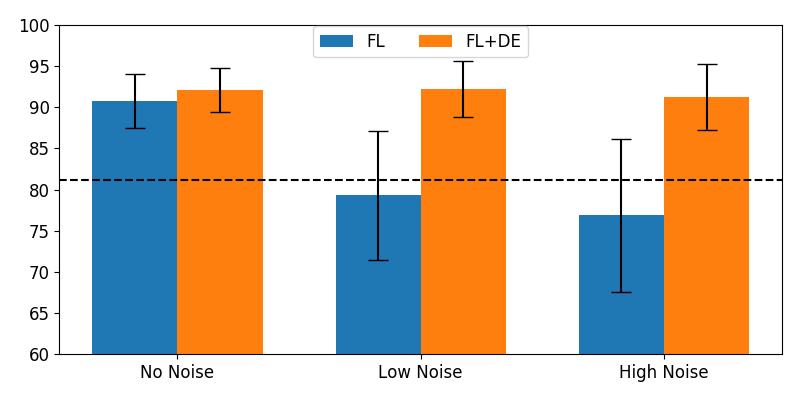}
}{
\caption{
Classifier accuracy on the spam dataset (higher is better).
FL and our system (FL+DE) are evaluated as gradient noise is increased
(for differential privacy).
The dashed horizontal line indicates domain-only baseline performance.
Error bars show performance variance across users.
}
\vspace{-0.2in}
\label{spam-b-results}
}
\end{floatrow}
\end{figure}

Test errors for the baseline, FL, FL+DE systems are provided in
Table~\ref{synthetic-table}. FL+DE provides the best results of any model,
and graceful degradation compared to differentially-private FL as the noise increases.
FL alone provides a lower error for both users if there are no privacy concerns, but as we
increase the noise we apply to the gradient, we observe a dramatic
increase in error.
FL+DE is more expressive than a single linear model - 
it learns a linear model and gating function for each user on top of the shared linear
model - so it is unsurprising that RMSE is lower when no noise is
added to the gradients. However, FL+DE does not suffer as much penalty
as FL when noise is added to the shared gradient updates.
In the worst case, the performance degrades only to the
baseline level (where each user has a linear model for its entire
dataset).

The second dataset is a real-world domain adaptation dataset for spam
detection, which was released as part of the ECML PKDD 2006 Discovery
Challenge \cite{bickel06}.  The task is to classify whether an email
in a user's inbox is spam, and personalizing the spam filtering for
each user. The amount of data available per user is limited, so it is
expected that collaboration can increase the quality of the
classifier. However, each user has a different inbox, so domain
adaptation is required. The dataset was originally designed to test
methods of unsupervised domain adaptation, but using the evaluation
dataset labels, which are now publicly available, we simulate $15$
users collaborating to build a spam classifier in a supervised setting. In this case, we
measure classifier accuracy, not prediction error. We use the dataset
from task $b$.  For each of our users, we train on $50$ labeled
examples, leaving $350$ examples for testing. The baseline
system trains one classifier for each user, using in-domain data only,
and we also train a collaborative FL model and finally FL+DE.

The results are illustrated in Figure~\ref{spam-b-results}. Once again,
FL+DE provides the best overall accuracy, and maintains its performance
as noise is added to provide differential privacy.
The accuracy of the baseline system is 81.2\% ($\pm$ 10.5\%), averaged
across users. In the absence of noise, FL achieves 90.8\% mean accuracy, and
FL+DE reaches 92.1\% accuracy. When we simulate low noise ($\sigma=0.5$),
FL classification accuracy drops to 79.3\%, but FL+DE achieves a minor improvement to
92.2\% mean accuracy. In high noise ($\sigma=1.0$), the FL system accuracy drops further to 76.9\%,
but the FL+DE accuracy drops only to 91.3\%.
On this dataset, we change parameters for low and high noise to $\sigma=0.5$ and $\sigma=1$,
because higher noise levels led the FL model to worse-than-random perfomance for some users,
while the FL+DE accuracy dropped to roughly the baseline accuracy.

\section{Related Work}
\label{sec:related}

Our core idea of maintaining two models per party is somewhat similar
to Daum{\'{e}} and Marcu's work~\cite{daume06}, where they propose
training three models, a ``source-domain'' model, a ``target-domain''
model, and a ``general'' model.  The composition of
these models during training and for inference is quite complex,
whereas our approach uses a simple weighted averaging algorithm.
Furthermore, their work targets two domains, whereas ours is more
suitable for FL and can target a multitude of domains, one for each participating
party.

Domain adaptation and federated learning have been studied in privacy-preserving and secure settings.
One line of work focuses on protecting privacy in a classic
domain adaptation setup~\cite{guo2018}, where a well-tuned model on a source domain is
adapted to perform better in a target domain with more limited data.
Another line of work focuses on secure federated learning~\cite{liu2018},
but uses additively homomorphic encryption to ensure privacy in a two-party
federated learning context. This is different from $\varepsilon$-differential privacy,
and does not maintain a collaborative general model. Each of these systems considers one part of
our set-up, but no prior work combines efforts of collaborative
learning combined with private domain adaptation.

Attention~\cite{bahdanau2014} has been used in a FL environment to weight updates
from different users~\cite{ji2018}, allowing users with extremely unusual
gradient updates to have a smaller disruptive effect on the general
FL model. While this generally improved perplexity on unseen data,
improving the general model, it does not allow users with unusual datasets
to improve prediction quality on their domain. 

The PATE architecture~\cite{papernot18} is yet another class of
distributed ML systems that uses privately trained models of
participating parties as sources for consensus-based labeling of data
for a new user to help it train its model on its private data.  The
models trained for individual users act as an ensemble of
``teachers'' for the new party that is training a new model for
itself.  The consensus based labeling provides the privacy guarantees
for each party.  This approach could be used to build the general model,
rather than differentially-private FL, but we have not yet tested its
effectiveness in conjunction with domain-adaptation techniques.

\section{Conclusion}

This work demonstrates that adding private, per-user domain adaptation to 
a collaborative model-building framework can increase accuracy for all users,
and is especially beneficial when privacy guarantees begin to diminish the
utility of the collaborative general model.

Our implementation of domain adaptation employs a mixture of experts, with each
user learning a domain expert model and a private gating mechanism.
This domain adaptation framework is another contribution
of our work, and allows us to train the entire model
with gradient descent. We demonstrate that it works well in practice on both
regression and classification tasks.

In future work, however, it may be practical to consider other mechanisms
for building a collaborative model (e.g., PATE), or alternative domain adaptation techniques (e.g., hypothesis
transfer learning).
We expect that the general setup of learning one collaborative generalist
and a private domain adaptation mechanism will be useful in many settings and
for many types of models.

\bibliographystyle{plain}
\bibliography{refs}

\end{document}